# The Free-Market Algorithm: Self-Organizing Optimization for Open-Ended Complex Systems

*Martin Jaraiz — Department of Electronics, University of Valladolid, Spain*
*March 2026*

**Abstract.** We introduce the Free-Market Algorithm (FMA), a novel metaheuristic optimization framework inspired by free-market economics. Unlike Genetic Algorithms, Particle Swarm Optimization, Ant Colony Optimization, and Simulated Annealing—all of which require a prescribed fitness function and search a fixed solution space—FMA uses distributed supply-and-demand dynamics in which fitness is *emergent*, the search space is *open-ended*, and solutions take the form of complete hierarchical pathway networks rather than single optima. Autonomous agents discover transformation rules, trade goods, open and close firms, and compete for demand, with no centralized controller. We present the formal algorithm, its three-layer architecture (universal market engine, pluggable domain rules, domain-specific observation), and compare it systematically with established metaheuristics. We further show that FMA provides a constructive mechanism for the abstract signatures of selection described by Assembly Theory. The algorithm has been validated in two unrelated domains: prebiotic chemistry—where it discovers all 12 feasible amino acids, all 5 nucleobases, and the complete formose sugar chain from bare atoms in under 5 minutes (Jaraiz, 2026b)—and macroeconomic forecasting—where it achieves Mean Absolute Error (MAE) of 0.42 percentage points for non-crisis GDP prediction with zero calibrated parameters, portable to 33 countries (Jaraiz, 2026a). These results suggest that Darwinian free-market optimization constitutes a universal mechanism for selection across complex systems.

**Contents**

# 1. Introduction

Complex systems resist forecasting because they violate the foundational assumptions of most computational methods: they have no fixed phase space, no representative agent, and no equilibrium to converge toward. An economy is not a system of equations to be solved but an evolving ecology of heterogeneous firms. A prebiotic chemical mixture is not a reaction network to be enumerated but an open-ended exploration of molecular space. In both cases, the system continually generates novelty that could not have been predicted from initial conditions.

Over the past fifty years, a rich family of *metaheuristic* algorithms has emerged, each drawing inspiration from a different natural process: Genetic Algorithms[1] from Darwinian evolution, Particle Swarm Optimization[2] from flocking behaviour, Ant Colony Optimization[3] from pheromone-guided foraging, and Simulated Annealing[4] from metallurgical cooling. These methods have been applied successfully to scheduling, routing, parameter estimation, and countless other domains.

Despite their diversity of metaphor, these algorithms share three structural assumptions that limit their applicability to open-ended complex systems:

1. **Prescribed fitness.** The quality of every candidate solution is evaluated by a function defined *a priori* by the designer. The algorithm does not decide what is valuable; it only searches for more of what it has been told to value.
2. **Fixed search space.** The set of possible solutions is defined before the run begins. A GA searches binary strings of length *n*; PSO moves particles through $\mathbb{R}^d$; ACO constructs paths on a given graph. No new dimensions or building blocks appear during execution.
3. **Centralized selection.** A global mechanism—tournament selection, global best update, pheromone reinforcement, Boltzmann acceptance—decides which solutions survive. Information about quality flows through a central bottleneck.

These assumptions are entirely reasonable when the problem is well-posed: when we know what we are looking for, know the space in which to look, and can afford a central arbiter of quality. But many of the most important problems in science and engineering violate all three assumptions. In prebiotic chemistry, we do not know which molecules are "fit"—fitness emerges from thermodynamic stability and mutual compatibility. In macroeconomics, GDP is not a fitness function to be maximized but an emergent aggregate of millions of interacting firms. In open-ended evolution, the search space itself grows as new building blocks are discovered.

**The economic insight.** Adam Smith observed in 1776 that individuals pursuing their own self-interest are "led by an invisible hand to promote an end which was no part of [their] intention"[9]. Hayek formalized this as a distributed information-processing mechanism: the price system aggregates knowledge that no central planner could collect[10]. We take this insight literally. The market is not a metaphor for optimization—it *is* an optimization algorithm, one whose key properties (emergent fitness, open-ended search, distributed selection) are precisely what existing metaheuristics lack.

In this paper we introduce the **Free-Market Algorithm (FMA)**, implemented as a Free-Market Agent-Based Model (FM-ABM), a metaheuristic that uses the market mechanism as a computational optimization framework. FMA's key innovations are:

- **Emergent fitness.** No objective function is prescribed. "Fitness" emerges from what the market demands and sustains.
- **Open-ended search space.** A persistent recipe book grows during execution; each discovery expands the space of possible future discoveries.
- **Distributed selection.** No central controller selects winners. Agents independently choose, compete, succeed or fail.
- **Hierarchical network solutions.** The output is not a single optimum but a complete production network with supply chains.
- **Persistent memory.** The recipe book is never cleared, enabling hierarchical discovery across arbitrary depth.

We present the formal specification, analyse its properties, compare it with established algorithms, demonstrate its connection to Assembly Theory[5], and summarize validated results in prebiotic chemistry and macroeconomic forecasting.

## 2. A Taxonomy of Complexity

To understand what FMA is designed for—and why existing metaheuristics fall short—we must first understand the landscape of complexity. Not all complex systems are alike, and different kinds of complexity demand different computational tools.

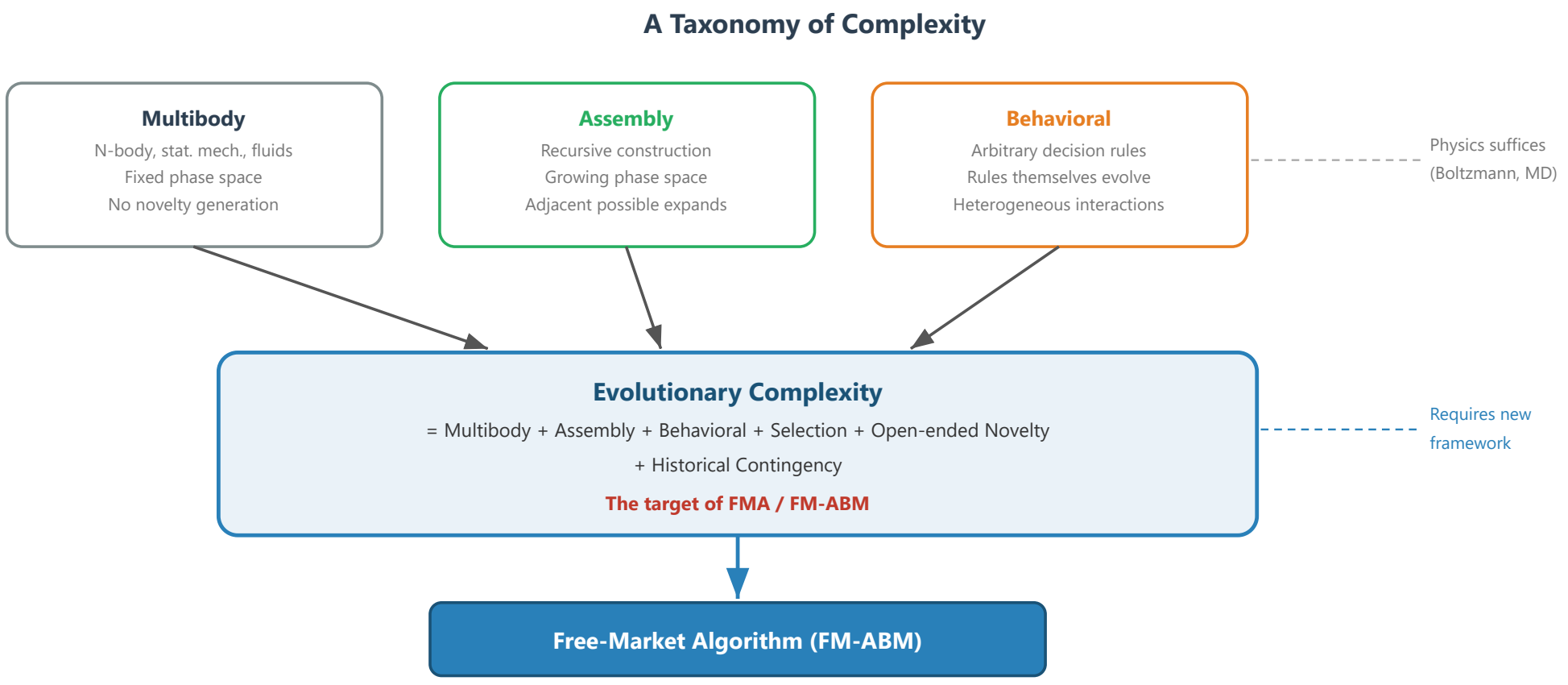

**Figure 1 | A taxonomy of complexity.** Multibody systems have fixed phase spaces and are well-served by physics. Assembly systems generate new possibilities through recursive construction. Behavioral systems feature agents with arbitrary evolving rules. Evolutionary complexity is the union of all three plus selection, open-ended novelty, and historical contingency—the target of the Free-Market Algorithm.

## 2.1 Multibody Complexity

The simplest form of complexity arises from many particles interacting through known forces: N-body gravitational systems, molecular dynamics, fluid turbulence. These systems are conceptually straightforward—the equations of motion are known, conservation laws constrain the dynamics, and Boltzmann statistics describe equilibrium. The difficulty is purely computational: the phase space is vast but *fixed*. No genuinely new objects appear; the set of possible states is determined at the outset. This is why statistical mechanics succeeds: it needs only sample a pre-existing landscape.

## 2.2 Assembly Complexity

A qualitatively different kind of complexity emerges when objects are built recursively from simpler ones: atoms combine into molecules, molecules into polymers, polymers into functional complexes. Assembly Theory (AT), formalized by Sharma et al.[5], provides a rigorous framework for measuring this complexity through two quantities: the *assembly index* (minimum recursive joining operations to construct an object from basic building blocks) and the *copy number* (how many identical copies exist in a sample). AT's central insight is that objects with both high assembly index and high copy number cannot arise from undirected physics alone—they require selection.

The critical distinction from multibody complexity is that the phase space *grows*. Each newly assembled object creates possibilities that did not exist before—what Kauffman[8] calls the "adjacent possible." A molecule that did not exist at step $t$ becomes a potential reactant at step $t+1$, opening pathways that were previously inaccessible.

### 2.3 Behavioral Complexity

A third form of complexity arises when agents follow arbitrary decision rules that need not correspond to any assembly operation. A bank does not *assemble* anything when it sets interest rates based on risk assessment. A government does not *construct* objects when it allocates spending based on political priorities. A foraging animal does not *join* building blocks when it selects a migration route. Yet these agents participate in systems every bit as complex as molecular evolution—their interactions generate emergent patterns (business cycles, ecosystem dynamics, social hierarchies) that no individual agent intends or controls.

The hallmark of behavioral complexity is that the rules themselves evolve: agents learn, adapt, imitate, and innovate. No single equation captures the system, because the system's dynamics depend on the heterogeneous, time-varying decision functions of its constituent agents.

### 2.4 Evolutionary Complexity

The most challenging systems combine all three forms above with three additional properties:

- **Selection:** some variants persist and proliferate while others disappear, but the selection criterion is not externally specified—it emerges from the system's dynamics.
- **Open-ended novelty:** genuinely new entities appear that could not have been predicted from initial conditions. The system continually expands its own state space.
- **Historical contingency:** what exists now depends on what existed before. The same initial conditions can produce radically different outcomes depending on the sequence of events.

These are the systems—economies, ecosystems, technological evolution, prebiotic chemistry—where forecasting has the greatest value and the least reliability. Traditional forecasting methods fail because they assume fixed phase spaces, representative agents, equilibrium convergence, and historically stable parameters. What is needed is a framework that generates novelty endogenously, accommodates heterogeneous agents with diverse and evolving rules, operates out of equilibrium, and produces selection without hand-designed fitness functions. The Free-Market Algorithm provides such a framework.

## 3. The Free-Market Algorithm

### 3.1 Design Principles

FMA is built on a single foundational principle: **strict separation of mechanism from domain rules**. The market mechanism—how agents trade, compete, enter and exit—is universal and identical across all applications. The domain-specific rules—how goods are created and whether transformations are feasible—are pluggable modules that encode domain knowledge without prescribing what the system should find. This separation ensures that selection emerges from market dynamics, not from the domain rules themselves.

Three economic insights, taken literally as computational mechanisms, underpin the design:

#### Adam Smith's Invisible Hand

Individual agents pursue their own interest without knowledge of the global outcome. Yet the collective result is a highly organized production network. In FMA, this is the selection mechanism itself.

#### Hayek's Knowledge Problem

No single entity possesses the information needed for optimal allocation. Knowledge is dispersed among agents. The trading mechanism aggregates distributed information, achieving coordination without central planning.

#### Schumpeter's Creative Destruction

Entry and exit of firms is a feature, not a bug. Firms that produce unwanted goods close, freeing agents to open new firms. This turnover balances exploration and exploitation.

**Clarification.** FMA is not a simulation of an economy. It is an *algorithm* that uses market dynamics as a computational mechanism, just as GA uses natural selection and ACO uses pheromone-guided foraging. The claim is not that markets are optimal in the normative sense, but that the market mechanism—distributed agents, local information, supply-and-demand equilibration—is a powerful optimization primitive that has been underexploited in the metaheuristic literature.

### 3.2 The Three-Layer Architecture

FMA achieves domain generality through a strict three-layer separation:

| Layer | Responsibility | Changes across domains? | Examples |
|---|---|---|---|
| **Layer 1: Market Engine** | Agent lifecycle, ring topology, trading, firm entry/exit, demand propagation, decay | No—identical in all domains | Ring neighbourhood, GFCF demand, patience threshold |
| **Layer 2: Domain Rules** | `DomainCombine()` and `DomainValidate()` —how goods are created and whether transformations are feasible | Yes—pluggable per domain | Chemistry: BDE + UFF energy. Economics: I–O coefficients. Robotics: task dependencies |
| **Layer 3: Observation** | Domain-specific metrics computed from the market state | Yes—read-only, no feedback | Chemistry: molecular census, assembly index. Economics: GDP, sectoral output |

This separation means that the market mechanism (Layer 1) is a *universal* optimization engine. To apply FMA to a new domain, one writes only the combination/validation rules (Layer 2) and the observation metrics (Layer 3). The trading, competition, discovery and demand mechanisms require no modification.

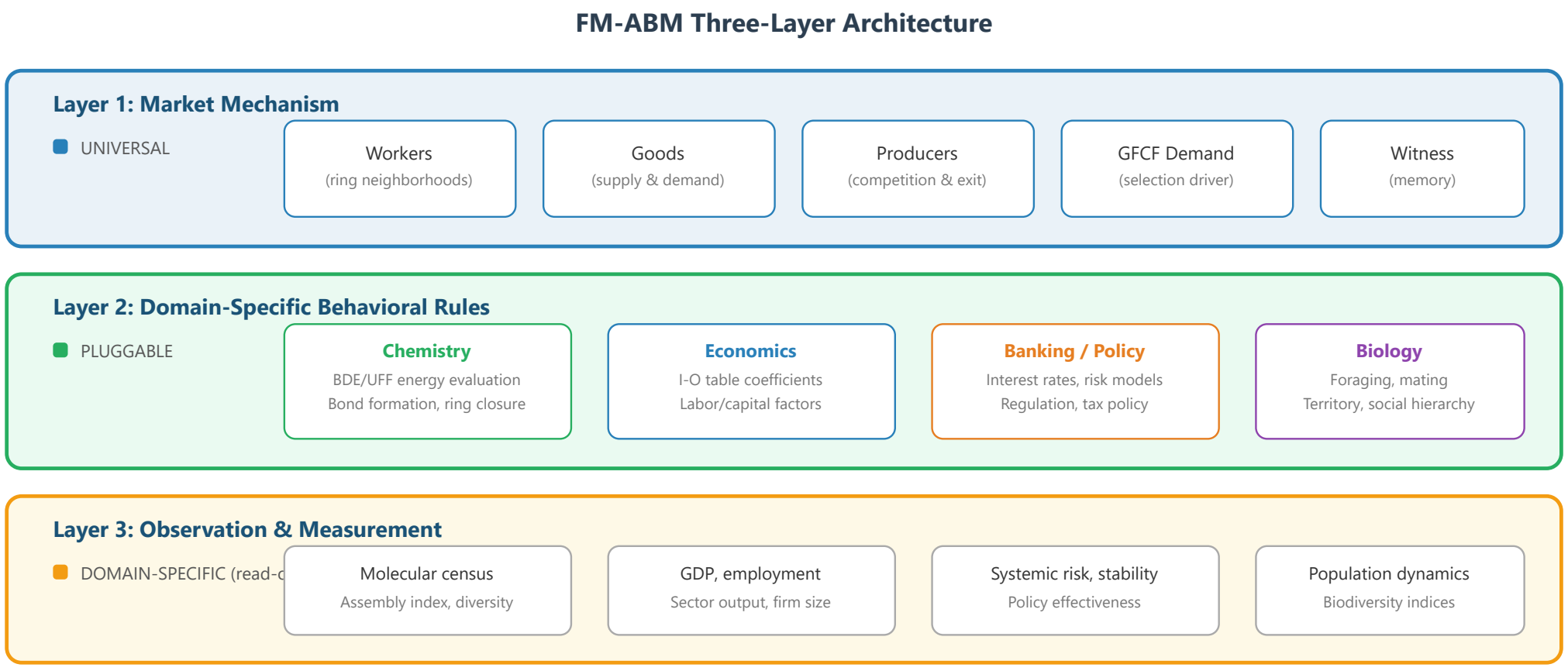

**Figure 2 | FM-ABM three-layer architecture.** Layer 1 (market mechanism) is universal and identical across all applications. Layer 2 (behavioral rules) is pluggable—different domains supply different rules for how agents interact and transform goods. Layer 3 (observation) extracts domain-specific measurements from the simulation state. Selection emerges from market dynamics (Layer 1), not from the domain rules (Layer 2).

## 3.3 The Agent's Activity Cycle

Each simulation step, every agent executes a five-phase economic cycle—the "Workshop Manager's Day":

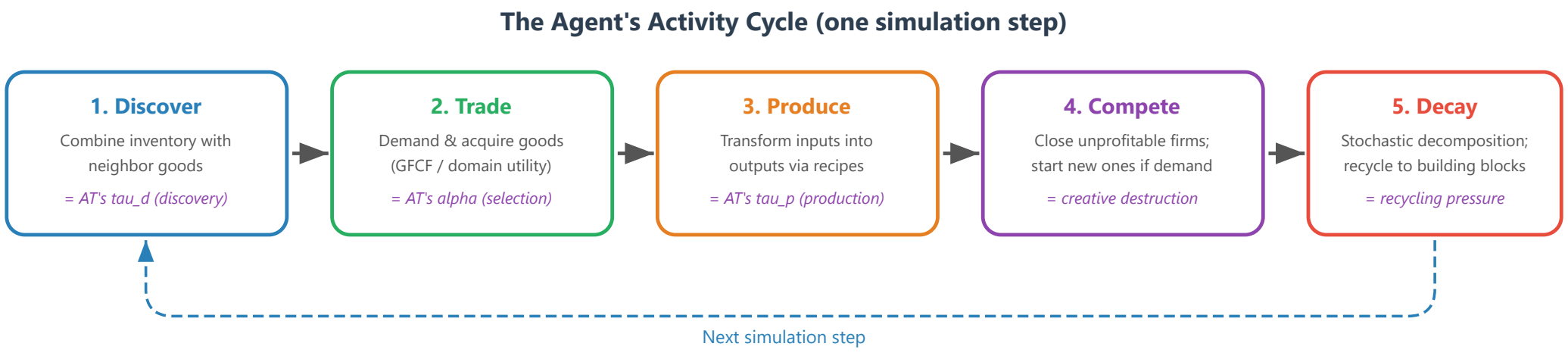

**Figure 3 | The agent's activity cycle.** Every agent executes five phases each simulation step. The cycle maps directly onto Assembly Theory's dynamical processes: Discover controls tau_d, Trade implements alpha, Produce controls tau_p, Compete provides creative destruction, and Decay recycles resources. No phase is designed *for* AT—the mapping is emergent.

**Phase 1: DISCOVER.** A bounded number of discovery attempts (the *budget*) are made per step. Each attempt picks a random agent, selects a good from its inventory, selects a good from a random neighbour's inventory, and asks the domain-specific combination function to propose a product. If the domain-specific validation function accepts the product (e.g., negative energy change in chemistry, feasible input–output coefficients in economics), a new recipe is added to the persistent recipe book, and the discovering agent receives a "witness sample" of the product.

**Phase 2: TRADE.** Each agent inspects the recipe book and expresses demand for the goods it prefers. Under GFCF demand, agents prefer the most complex (highest-depth) goods available, creating a demand cascade: demand for a depth-3 molecule generates derived demand for its depth-2 inputs, which generates demand for depth-1 inputs, and so on. This is the "invisible hand"—nobody plans the supply chain; it emerges from individual preferences. Agents that have opened a producer attempt to buy the inputs they need from neighbours.

**Phase 3: PRODUCE.** If all inputs are secured, production occurs. Trade is strictly local: an agent can only buy from its *k* ring neighbours, not from arbitrary agents. This locality creates spatial structure and prevents trivially optimal allocation.

**Phase 4: COMPETE.** Firms that cannot sell their output accumulate losses. After a patience threshold, they close. Idle agents survey unmet demand in their neighbourhood and may open a new firm to fill the gap. This is Schumpeter's *creative destruction*: the entry and exit of firms is the algorithm's mechanism for balancing exploration and exploitation.

**Phase 5: DECAY.** Goods in inventories stochastically decompose into their component parts, with a decay rate proportional to their complexity. Decay serves two purposes: it recycles material back into the pool of simple building blocks, preventing resource lock-up; and it acts as a selection pressure against goods that are merely complex without being demanded (demanded goods are continually replenished by production).

### 3.4 Formal Pseudocode

```
 FMA(N agents, domain_rules, max_iterations)
━━━━━━━━━━━━━━━━━━━━━━━━━━━━━━━━━━━━━━━
Initialize: N agents on ring, each with one primary good
RecipeBook ← ∅

for t = 1 to max_iterations:

  // Phase 1: DISCOVER (exploration)
  for budget times:
    agent_i ← random worker
    good_A  ← from agent_i.inventory
    good_B  ← from random neighbor's inventory
```

```
    product ← DomainCombine(good_A, good_B)
    if DomainValidate(product):           // e.g., ΔE < 0
      RecipeBook ← RecipeBook ∪ {(A, B) → product}
      agent_i.inventory += product        // witness sample

  // Phase 2: DEMAND (selection signal)
  for each agent:
    wishlist ← SelectPreferred(RecipeBook)   // GFCF
    RecordDemand(wishlist)

  // Phase 3: TRADE & PRODUCE (distributed exchange)
  for each agent with producer:
    for each input needed:
      seller ← FindInNeighborhood(input)
      if seller: Transfer(input, seller → agent)
    if all inputs available:
      output ← Produce(recipe)
      agent.inventory += output

  // Phase 4: COMPETE (market selection)
  for each agent with producer:
    if unprofitable for too long:
      CloseProducer()                     // creative destruction
  for each idle agent:
    if unmet demand exists:
      StartProducer(demanded_recipe)      // entrepreneurship

  // Phase 5: DECAY (recycling)
  for each good in all inventories:
    if random() < DecayRate(good):
      Decompose(good) → components       // returns to pool

return RecipeBook, all inventories, production_network
```

### 3.5 Key Properties

**Property 1: Emergent fitness.** FMA has no prescribed objective function. What counts as "fit" is determined by the interplay of discovery, demand and competition. A molecule that is thermodynamically stable, demanded by consumers, and efficiently producible from available inputs will be sustained by the market. Fitness is a *systems property*, not a scalar function.

**Property 2: Open-ended search space.** The recipe book starts empty and grows monotonically. Each new recipe potentially enables the discovery of higher-depth products that were previously unreachable. This is a computational realization of Kauffman's "adjacent possible"[8]: the set of things that can be explored expands with each exploration. No existing metaheuristic has this property.

**Property 3: Distributed selection.** Selection in FMA is fully decentralized. No global fitness ranking exists. Each agent makes local decisions (what to demand, what to produce, when to close). The emergent behaviour—complex production networks, supply chains, specialization—arises from these local interactions. This is the computational analogue of Hayek's insight that the price system is a distributed computer[10].

**Property 4: Persistent memory.** The recipe book is never cleared. Unlike GA (where each generation is memoryless) or SA (which has no memory), FMA accumulates knowledge across all iterations. This memory is essential for hierarchical discovery: a depth-3 product can only be discovered after its depth-2 inputs have been discovered, which requires the depth-1 recipes to persist.

**Property 5: Two timescales.** FMA's behaviour is governed by the ratio of two rates: the discovery rate $\tau_d$ (controlled by the budget parameter) and the production rate $\tau_p$ (controlled by the number of agents and neighbourhood size). When $\tau_d >> \tau_p$, the system explores rapidly but produces little. When $\tau_p >> \tau_d$, the system exploits known recipes efficiently but discovers nothing new. The balance between these timescales is FMA's analogue of the exploration/exploitation trade-off, but it emerges naturally from the market dynamics rather than being imposed by a schedule.

**Property 6: Hierarchical network solutions.** FMA's output is not a single "best" solution but a complete production network: a directed acyclic graph of recipes, supply chains, inventories and active firms. This network encodes not just *what* was found optimal but *how* it is produced, from raw materials through intermediates to final products. For many applications—chemical synthesis, manufacturing, logistics—the pathway is as valuable as the destination.

### 3.6 Comparison with Existing Metaheuristics

Table 1 provides a systematic comparison of FMA with four established metaheuristics across nine structural properties.

| Property | GA[1] | PSO[2] | ACO[3] | SA[4] | FMA |
|---|---|---|---|---|---|
| **Fitness function** | Prescribed | Prescribed | Prescribed | Prescribed | **Emergent** |
| **Search space** | Fixed | Fixed | Fixed (graph) | Fixed | **Open-ended** |
| **Selection mechanism** | Centralized | Global best | Pheromone | Boltzmann | **Distributed** |
| **Memory** | None | Personal + global best | Pheromone trails | None | **Recipe book + inventories** |
| **Solution type** | Single optimum | Single optimum | Path | Single optimum | **Network of pathways** |
| **Population structure** | Flat (panmictic) | Flat (swarm) | Graph | Single agent | **Ring + firms** |
| **Hierarchy** | None | None | None | None | **Supply chains** |
| **Discovery mechanism** | Crossover + mutation | Velocity update | Probabilistic construction | Random perturbation | **Combination + validation** |
| **Exploration / exploitation** | Mutation rate | Inertia weight | Evaporation rate | Temperature schedule | **$\tau_d$ / $\tau_p$ balance** |

**Table 1 | Systematic comparison of FMA with established metaheuristics.** FMA is the only algorithm with emergent fitness, open-ended search, distributed selection, persistent memory, and hierarchical solutions. These properties are not incremental improvements but qualitative differences that make FMA applicable to problem classes that existing algorithms cannot address.

In summary: when you know what you're looking for and where to look, use GA. When you want to discover *what exists* and *how it can be made*, use FMA.

## 4. Connection to Assembly Theory

Assembly Theory (AT), introduced by Sharma et al.[5], provides a mathematical framework for quantifying selection and evolution. AT defines two key observables: the *assembly index* (the minimum number of joining operations required to construct an object from basic building blocks) and the *copy number* (how many copies of that object are observed). AT's central claim is that objects with high assembly index and high copy number are signatures of selection—they cannot arise by chance and must be the product of an information-generating process.

FMA provides a *constructive mechanism* for the phenomena AT describes abstractly. The mapping is precise and unforced:

### 4.1 Mapping AT Concepts to FMA

| Assembly Theory Concept | FMA Implementation |
|---|---|
| **Assembly index** | Recipe depth in the recipe book. A depth-3 molecule requires three sequential joining operations, each corresponding to a recipe |
| **Copy number** | Census count—the number of copies of a molecular formula present in all agent inventories at a given time step |
| **α (selectivity parameter)** | GFCF demand mechanism. When α < 1 (directed demand), agents prefer complex goods, creating the selective amplification that AT predicts. In FMA, this emerges from economic dynamics rather than being set as an input parameter |
| **$\tau_d$ / $\tau_p$ phase diagram** | Discovery budget parameter. AT predicts three regimes depending on the ratio of discovery to production rates. FMA reproduces all three from a single tunable parameter |
| **Non-monotonic copy** | Under GFCF demand (α < 1), FMA shows peaked copy-number distributions at intermediate depth—AT's signature of selection. |
| **distribution** | Under blind demand (α = 1), the distribution is monotonically decreasing—AT's signature of random assembly |
| **Assembly space** | The recipe book *is* the assembly space—all discovered pathways for all observed objects, progressively explored through thermodynamically grounded discovery |

## 4.2 Three Dynamical Regimes

AT predicts three dynamical regimes based on the ratio of production timescale $\tau_p$ to discovery timescale $\tau_d$. FMA reproduces all three by varying a single parameter: `discoveryBudget` .

**Budget = 1: Stagnation**

$\tau_p << \tau_d$

**132** species, **6,817** objects, max depth **12**, *A* = **987**

Few species discovered but heavily amplified. Production outpaces exploration. Selection signature visible: non-monotonic copy distribution emerges.

**Budget = 10: Selection Sweet Spot**

$\tau_d \approx \tau_p$

**784** species, **14,138** objects, max depth **12**, *A* = **377**

Balanced exploration and exploitation. Maximal diversity with sustained production. The optimal operating point.

**Budget = 50: Tar**

$\tau_p >> \tau_d$

**22** species, **70** objects, max depth **5**, *A* = **10**

Combinatorial explosion—everything is discovered, nothing persists. Resources spread too thin for any production chain to sustain itself.

The non-monotonic copy-number distribution at Budget = 1 is particularly striking. Copy numbers decline from depth 1 (530) through depth 5 (7), then *rise* at depths 6–9 (29, 14, 38, 26). Copies at depth 8 (avg = 38) exceed copies at depth 4 (avg = 11). This 3.5x amplification at higher assembly index is AT's "smoking gun" for selection—impossible under undirected production—and it emerges spontaneously through market dynamics.

### 4.3 FMA as the Mechanism AT Describes

The relationship between AT and FMA is complementary: AT asks "Given an observed distribution of objects, was selection at work?" FMA answers the constructive question: "Given a set of building blocks and combination rules, what does selection produce?" Together, they form complementary perspectives on the same phenomenon.

**Key result.** FMA with GFCF demand produces the non-monotonic copy distribution at high assembly depth that Assembly Theory identifies as the universal signature of selection. This is not imposed—it *emerges* from the market dynamics. FMA thus provides the *mechanism* that AT describes: selection acts through the invisible hand of supply and demand, not through an external fitness function.

| Dimension | Assembly Theory | FMA / FM-ABM |
|---|---|---|
| **Core function** | Measures assembly (detection) | Generates assembly (mechanism) |
| **Selection** | "Put in by hand" via α | Emerges from market dynamics |
| **Temporal direction** | Post-hoc measurement | Forward prediction |
| **Physics grounding** | Abstract (any joining operation) | Concrete (BDE/UFF/Evans–Polanyi) |
| **Novelty detection** | "Physics cannot distinguish" | Neighbor interaction + consumption |
| **Formation histories** | Theoretical construct | Empirical recipe book |

In AT's own words: "We are putting selection in by hand in our examples to demonstrate foundational principles."[5] FMA demonstrates that selection need not be put in by hand—it arises naturally from economic dynamics applied to domain-specific physics. Assembly Theory is the thermometer. FM-ABM is the weather model.

## 5. Application I: Prebiotic Chemistry

In the prebiotic chemistry application, FMA's Layer 2 is instantiated with Bond Dissociation Energies (BDE) and Universal Force Field (UFF) strain evaluation. The domain combination function proposes a covalent bond between two molecular fragments; the domain validation function accepts the product only if the total energy change ΔE is negative (exothermic). Evans–Polanyi kinetic barriers gate reaction rates through temperature-dependent acceptance probabilities.

**Results summary.** Starting from 200 C + 500 H + 100 O + 100 N atoms (900 total) with no target molecules specified, FMA discovered:

- All **12 feasible amino acids** from C, H, O, N—including the same "early 10" identified by Miller–Urey experiments, meteorite analysis, and phylogenetic reconstruction
- All **5 nucleobases**—the letters of RNA and DNA (adenine, guanine, cytosine, uracil, thymine)
- The **complete formose sugar chain** from formaldehyde to ribose to glucose
- **Krebs cycle intermediates** (pyruvate, succinate, malate)
- The first **tripeptide** (Gly-Gly-Gly)
- Up to **260 independent synthesis routes** for a single amino acid

Amino acid formulas appeared **14 times earlier** than random chance would predict: by step 50, 7 of 12 amino acid formulas were among the 213 species discovered (3.3%), versus 0.24% at full exploration. This quantifies the thermodynamic inevitability hypothesis—life's building blocks are not accidents but thermodynamic attractors in assembly space.

The simulation runs on a standard laptop in under 5 minutes—4 to 6 orders of magnitude cheaper than molecular dynamics or DFT methods for the same chemical space. Full details of the chemistry application, including reactor scaling, thermodynamic evolution, pathway analysis, and 8 independent synthesis routes to histamine, are reported in the companion paper (Jaraiz, 2026b)[12].

## 6. Application II: Macroeconomic Forecasting

In the macroeconomics application, FMA's Layer 2 is instantiated with input–output (I–O) coefficients from the FIGARO database. Each agent represents a firm in one of 64 economic sectors. The domain combination function uses the I–O table to determine how much of each input sector a firm needs to produce one unit of its output sector. The domain validation function checks whether the firm can secure all required inputs from its neighbourhood.

The forecasting protocol is remarkable in its simplicity: one input–output table for year $N$ produces an FM-ABM forecast of GDP for year $N+1$. No time series. No estimated parameters. No external forecasts. No behavioral assumptions about expectations, policy responses, or market sentiment. The economy self-organizes from its structural information alone.

| Metric | FM-ABM | Best Professional Forecaster | European Commission |
|---|---|---|---|
| **Austrian 9-year panel MAE** | **1.22 pp** | 0.91 pp (WIFO) | 1.15 pp |
| **Non-crisis years MAE** | **0.42 pp** | 0.48 pp (WIFO) | 0.72 pp |
| **Cross-country convergence** | **33 of 37 countries** | Country-specific | Country-specific |
| **Parameter calibration** | **None** | Expert judgment | Econometric models |

**Table 2 | GDP forecasting performance.** FM-ABM achieves MAE comparable to professional forecasters for non-crisis years, with zero calibrated parameters. pp = percentage points.

**Key insight.** The same algorithm that discovers amino acids from bare atoms also forecasts GDP from input–output tables. Only the Layer 2 domain rules changed—the market engine (Layer 1) was identical. This is the strongest evidence for FMA's universality: comparable institutional-quality accuracy in a completely unrelated domain. Full details are reported in the companion paper (Jaraiz, 2026a)[11].

## 7. Application III: Synthesis Route Planning

FMA's ability to discover multiple independent synthesis routes to a single target molecule, with full energy diagrams and temperature-dependent accessibility, makes it a practical tool for synthesis pathway design. We illustrate with histamine ($C_5H_9N_3$), a molecule of pharmaceutical importance.

**T = 300 K (27 °C)**

**5 independent routes** discovered, all at depth ≥ 4. Kinetic barriers restrict pathways to the most thermodynamically favorable multi-step sequences. Each route includes a complete supply chain tree from primary atoms through intermediates.

**T = 400 K (127 °C)**

**7 routes** discovered, including a **depth-3 shortcut** unlocked by the higher thermal energy. Practical implication: "raise temperature to 127 °C to access a shorter synthesis pathway." This is actionable information for laboratory chemists.

Each route comes with: (1) a complete energy profile showing ΔE at each step, (2) a supply chain tree tracing all inputs back to primary atoms, (3) a multi-criteria ranking (energy efficiency, depth, input availability), and (4) the temperature at which the route becomes accessible. This level of detail—multiple ranked routes with full provenance—is directly useful for retrosynthetic analysis and goes beyond what traditional retrosynthesis tools provide, as it discovers routes from first principles rather than from template databases.

The same framework has been applied to robot task sequencing (Jaraiz, 2020)[6], where FMA discovered feasible parallel execution schedules for multi-robot assembly. The emergent schedule matched or exceeded hand-crafted heuristic planners, with the advantage of automatic adaptation when task dependencies change.

## 8. Why FMA Works: The Economics of Exploration

A natural question arises: why should economic dynamics work as an optimization mechanism for chemistry or any other domain? The answer lies in the structural isomorphism between market optimization and the exploration/exploitation trade-off that all complex systems must resolve.

FMA's five selection mechanisms work in concert to navigate this trade-off:

**Demand**

Agents want complex/valuable goods. This pulls production toward complexity—no one needs to specify *which* goods are "fit."

**Competition**

Multiple producers compete for the same inputs. Efficient ones survive, inefficient ones close. Natural selection in economic form.

**Scarcity**

Finite resources force choices. Not everything can be produced simultaneously. This creates selective pressure driving specialization.

**Decay**

Unused products decompose. Their constituent resources are recycled into better uses. This clears the population of unsuccessful variants.

**Critical Producers (Witnesses)**

The last producer of a recipe is protected. This preserves rare knowledge—like state-owned enterprises providing essential services not covered by the private sector. In our simulations this protection is partly necessitated by the small number of atoms (~900) compared to the astronomical number of atoms in the prebiotic real world, where rare reactions would occur by sheer statistical weight.

**The supplier memory experiment.** A revealing experiment illustrates how FMA naturally adapts to domain requirements. In the economics application, agents remember their suppliers (exploitation dominates)—this reflects the institutional stability of real economic networks. In the chemistry application, agents do *not* remember suppliers—they react with a neighbor if that lowers the energy (exploration dominates)—this reflects the combinatorial nature of chemical discovery, where new combinations are continuously explored. The same algorithm, through its pluggable behavioral rules, naturally shifts the exploration/exploitation balance to match domain requirements. This is not a parameter to be tuned; it emerges from the domain rules themselves.

**Darwinian correspondence.** The Lewontin conditions for evolution by natural selection are satisfied naturally by market dynamics: **Variation** = discovery (combining existing things in new ways). **Selection** = market forces (demand, competition, scarcity). **Inheritance** = recipes persist across generations (simulation steps). FMA is not merely analogous to natural selection; it satisfies the formal requirements for a Darwinian process.

## 9. Discussion

### 9.1 FMA vs. GA: A Deeper Comparison

FMA and GA are both population-based metaheuristics inspired by natural processes, but they occupy fundamentally different niches in the algorithm design space:

| Criterion | Use GA when... | Use FMA when... |
|---|---|---|
| **Fitness function** | Known and computable | Unknown, emergent, or multi-dimensional |
| **Search space** | Fixed and well-defined | Open-ended or growing |
| **Desired output** | Single optimum or Pareto front | Complete pathway network with supply chains |
| **Solution structure** | Flat (no hierarchy) | Hierarchical (intermediates matter) |
| **Problem type** | Parameter optimization, scheduling, routing | Discovery, synthesis planning, network emergence |
| **Memory requirement** | None (generational replacement) | Essential (recipe book enables hierarchical discovery) |
| **Computational budget** | Tight (GA is cheaper per iteration) | Moderate (FMA needs local interactions at $O(N \cdot k)$) |

The deepest difference is philosophical. GA asks: "Given a fitness landscape, where is the peak?" FMA asks: "Given a set of building blocks and rules, what landscape emerges?" GA optimizes within a known space; FMA discovers the space itself.

### 9.2 Universality

The strongest evidence for FMA's universality is that the identical Layer 1 market engine—ring neighbourhoods, demand propagation, firm entry/exit, trading, decay—produced self-organized production networks in three fundamentally different domains:

- **Prebiotic chemistry:** Discovers amino acid synthesis from bare atoms in minutes. All 12 feasible amino acids, all 5 nucleobases, sugars, Krebs intermediates.
- **Macroeconomic forecasting:** Predicts GDP from input–output tables with MAE 0.42 pp for non-crisis years across 33 countries.

- **Robot task sequencing:** Discovers feasible parallel execution schedules for multi-robot assembly (Jaraiz, 2020)[6].

In each case, the emergent production networks matched empirical observations without domain-specific tuning of the market mechanism. Only the domain rules (Layer 2) and observation metrics (Layer 3) changed. This suggests that free-market dynamics may constitute a fundamental mechanism for selection across complex systems—a computational realization of the universal selection that Assembly Theory describes abstractly.

A deeper implication deserves attention. Assembly Theory introduces an intrinsic "assembly time" that ticks at each object being constructed—an *event-driven* clock rather than the continuous time of conventional physics. Our FM-ABM operates in precisely this mode: progress is measured in massively parallel assembly steps (discoveries, productions, trades), not in wall-clock seconds. This event-driven perspective resonates with several foundational programs in physics: causal set theory (Bombelli *et al.*, 1987), where spacetime geometry emerges from discrete causal events; relational quantum mechanics (Rovelli, 1996), where physical reality consists of interaction events rather than persistent states; the Wheeler–DeWitt equation (DeWitt, 1967), which contains no time variable at all; and constructor theory (Deutsch, 2013), which reformulates physics in terms of which transformations are possible rather than trajectories through time. If selection through market-like mechanisms operates from human economic activities all the way down to the molecular level (as our results across both domains suggest), it is natural to ask whether analogous event-driven assembly processes might also govern the construction of even more fundamental structures—from elementary particles to spacetime itself. The success of FMA at both the chemical and macroeconomic scales hints that Darwinian market dynamics may not merely *describe* selection but may reflect a deeper organizational principle that lead to the unfolding of Nature itself.

## 9.3 Limitations

We identify four limitations of the current FMA framework:

1. **Structural isomers (chemistry).** FMA operates on molecular formulas, correctly identifying compositions (e.g., $C_2H_5NO_2$ for glycine) but not always resolving the correct 3D structure among isomers. UFF validation eliminates geometrically strained structures, but the resulting isomer may differ from the biologically relevant one. This serves as a starting point for *ab initio* detailed determination of preferred isomers.

2. **Positive growth bias (economics).** The household wealth-consumption feedback loop produces systematic positive bias for economies with persistent low growth (France: MAE 2.76 pp) or high export dependency (Germany: MAE 3.68 pp)—a limitation of the *data* (Layer 2), not the *framework* (Layer 1). Multi-country network simulations using FIGARO's bilateral inter-country tables represent the natural extension.
3. **Computational cost per iteration.** Each FMA step involves N agents interacting with *k* neighbours, giving O(N·k) cost per step. This is more expensive than GA's O(N) per generation, though FMA's steps are richer in information content.
4. **Stochastic variability.** Like all metaheuristics, FMA is stochastic. Different random seeds produce different recipe books and production networks. The *qualitative* outcomes are robust (the same amino acids emerge across seeds), but quantitative metrics (exact copy counts, number of routes) vary.

## 10. Conclusions

The Free-Market Algorithm introduces a new class of metaheuristic optimization that fills a gap no existing algorithm addresses: optimization with *emergent fitness* in an *open-ended search space*, producing *hierarchical network solutions* through *distributed selection*. Its key innovation is the recognition that the market mechanism—agents trading, firms competing, demand cascading through supply chains—is not merely a metaphor for optimization but a powerful computational primitive in its own right.

The algorithm's contributions can be summarized as five innovations:

a. **No prescribed fitness function.** "Fitness" emerges from the interplay of discovery, demand, competition, scarcity, and decay. The algorithm does not need to be told what is valuable; it discovers value through market dynamics.
b. **Simultaneous discovery of solutions and search space.** The recipe book starts empty and grows monotonically. FMA discovers not just answers but the questions themselves—the space of possibilities that could not have been enumerated in advance.
c. **Natural exploration/exploitation balance.** The ratio $\tau_d/\tau_p$ self-regulates through market dynamics. No temperature schedule, no mutation rate adaptation, no evaporation tuning. The market finds its own balance.

d. **Persistent memory through recipe books and producer directories.** Knowledge accumulates across all iterations, enabling hierarchical discovery of arbitrary depth. This is essential for problems where solutions are built from solutions to sub-problems.
e. **Quantitative validation in two unrelated domains.** Chemistry: all 12 amino acids, 5 nucleobases, complete sugar chains from bare atoms. Economics: MAE 0.42 pp GDP forecasting across 33 countries with zero calibrated parameters. The identical Layer 1 engine produced both results.

The connection to Assembly Theory provides theoretical grounding: FMA generates the non-monotonic copy distributions at high assembly depth that AT identifies as the universal signature of selection. Where AT measures, FMA generates. Where AT describes, FMA constructs. Together they establish that Darwinian economic optimization is a viable—and perhaps fundamental—mechanism for the emergence of complexity.

We propose FMA as a general-purpose tool for any domain where the goal is not to find a single optimum but to discover the *landscape* of feasible solutions and their formation pathways. The three-layer architecture makes adoption straightforward: write a combination function, write a validation function, and let the market discover what is possible.

---

## References

1. Holland, J.H. *Adaptation in Natural and Artificial Systems*. University of Michigan Press (1975).
2. Kennedy, J. & Eberhart, R. Particle swarm optimization. *Proc. IEEE International Conference on Neural Networks*, 1942–1948 (1995).
3. Dorigo, M., Maniezzo, V. & Colorni, A. Ant system: optimization by a colony of cooperating agents. *IEEE Transactions on Systems, Man, and Cybernetics—Part B* **26**, 29–41 (1996).
4. Kirkpatrick, S., Gelatt, C.D. & Vecchi, M.P. Optimization by simulated annealing. *Science* **220**, 671–680 (1983).
5. Sharma, A., Czegel, D., Lachmann, M., Kempes, C.P., Walker, S.I. & Cronin, L. Assembly theory explains and quantifies selection and evolution. *Nature* **622**, 321–328 (2023).
6. Jaraiz, M. Ants, robots, humans: a self-organizing, complex systems modeling approach. *arXiv:2009.10823* (2020).
7. Nelson, R.R. & Winter, S.G. *An Evolutionary Theory of Economic Change*. Harvard University Press (1982).

8. Kauffman, S.A. *The Origins of Order: Self-Organization and Selection in Evolution*. Oxford University Press (1993).
9. Smith, A. *An Inquiry into the Nature and Causes of the Wealth of Nations* (1776).
10. Hayek, F.A. The use of knowledge in society. *American Economic Review* **35**, 519–530 (1945).
11. Jaraiz, M. Macroeconomic Forecasting from Input–Output Tables Alone: A Darwinian Agent-Based Approach with FIGARO Data. *arXiv:2603.12412* (2026a).
12. Jaraiz, M. Prebiotic Chemistry from First Principles: A Free-Market Agent-Based Model Discovers Life's Molecular Toolkit. *arXiv* (2026b, pending).
13. Miller, S.L. A production of amino acids under possible primitive Earth conditions. *Science* **117**, 528–529 (1953).
14. Bombelli, L., Lee, J., Meyer, D. & Sorkin, R.D. Space-time as a causal set. *Physical Review Letters* **59**, 521–524 (1987).
15. Rovelli, C. Relational quantum mechanics. *International Journal of Theoretical Physics* **35**, 1637–1678 (1996).
16. DeWitt, B.S. Quantum theory of gravity. I. The canonical theory. *Physical Review* **160**, 1113–1148 (1967).
17. Deutsch, D. Constructor theory. *Synthese* **190**, 4331–4359 (2013).

**AI Assistance Statement.** Claude (Anthropic) contributed to the implementation of the author's driving ideas, analysis of results, and drafting of this manuscript. The FMA concept, three-layer architecture, the insight connecting free-market dynamics to algorithmic optimization, and the connection to Assembly Theory are the author's original contributions developed over the course of the DEPLOYERS project (2020–2026). All scientific claims were verified by the author.

**Author's Disclosure.** For this report I provided the title, asked Claude to generate it and suggested only minor changes. All the chemistry-related work, from scratch, took "us" less than a week. I had worked on computational chemistry for several years.

March 2026